\pdfoutput=1

\documentclass[11pt]{article}

\usepackage{acl}

\usepackage{microtype}
\usepackage{times}
\usepackage{latexsym}

\usepackage[T1]{fontenc}

\usepackage[utf8]{inputenc}

\usepackage{microtype}

\title{Towards a general purpose machine translation system for Sranantongo}

\author{Just Zwennicker$^{\star\dagger}$ \quad David Stap\\
        Language Technology Lab\\ University of Amsterdam}

\begin{document}
\maketitle

\vspace{-10cm}
\begin{abstract}
Machine translation for Sranantongo (Sranan, srn), a low-resource Creole language spoken predominantly in Surinam, is virgin territory. In this study we create a general purpose machine translation system for srn. In order to facilitate this research, we introduce the SRNcorpus, a collection of parallel Dutch (nl) to srn and monolingual srn data. We experiment with a wide range of proven machine translation methods. Our results demonstrate a strong baseline machine translation system for srn.
\end{abstract}

\section{Introduction}\vspace{-0.15cm}
The official language in Surinam is Dutch, however the language that you will most likely hear on the streets of Paramaribo, its capital city, is Sranan. It is a lingua franca that finds its origins in the 17th century when the transatlantic slave trade brought people of different cultures and languages together. This Creole language broke down the language barrier between the different social groups.

There are approximately 600k speakers of Sranan worldwide \cite{srananpos_2021}, most of which live in Surinam, and in the Netherlands, their former colonizer. Around 1975, when Surinam became independent, a large group of people emigrated from Surinam to the Netherlands. That so called first generation is mostly fluent in Sranan, but this is often not the case for the second and third generation. A machine translation (MT) system for nl/srn could assist with (re)learning the language \cite{Lent2022} and facilitate familiarizing themselves with their cultural background.

Although Sranan is the second largest language spoken in Surinam there are relatively few written sources available. This could in part be explained by the stigmatization of the language, as is the case with many Creole languages \cite{Lent2022}. 

\vspace{-0.2cm}\section{Method, setup and results}\vspace{-0.15cm}
\textbf{SRNcorpus} We have collected data from various domains to create the SRNcorpus. It consists of parallel nl-srn as well as monolingual srn data. The largest data source contains religious data, in particular the Jehova Witness bible translations \citep{Agic2020}\footnote{Unfortunately the JW300 became unavailable during our research, reportedly due to copyright issues. This limited the options for our multilingual setup, as we were not able to secure data for other language pairs containing Sranan}. Furthermore we scraped around 3k non-religious, parallel sentences from an online Sranan dictionary\footnote{\href{http://suriname-languages.sil.org/Sranan/National/SrananNLDictIndex.html}{SIL parallel sentences}}. For numerous words listed in the dictionary, it contains an example sentence showing how it could be used, along with the Dutch translation of that sentence. This website also contained links  to various (children) stories\footnote{\href{http://suriname-languages.sil.org/Sranan/National/SrananNLLLIndex.html}{SIL stories}} in Sranan, yielding over 6.5k monolingual sentences. Lastly the SRNcorpus contains 5 smaller parallel sources from various domains. See Table \ref{tab:SRNcorpus} for a detailed description of its contents. 

While religious data is over-represented, our goal is to create a general purpose translation system. We therefore create non-religious validation and test sets, by sampling two times 256 parallel sentences from the non-religious sources within SRNcorpus to create test and validation sets.\\
\textbf{Other data} In addition to SRNcorpus, we used the following parallel data for our multilingual and transfer learning experiments: Wikimatrix \cite{schwenk-etal-2021-wikimatrix} (nl-English (en) 3.3M; nl-Portugese (pt) 1.2M), TED2020 \cite{reimers-2020-multilingual-sentence-bert} (nl-en 317K; nl-pt 260K) and Europarl \cite{koehn-2005-europarl} (nl-en 2M; nl-pt 1.9M). We used pt since srn inherited features from this language \cite{Seuren1981}, and similarity between parent and child languages can stimulate transfer \cite{Johnson2017}.

\begin{table}[h!]
  \scriptsize
  \centering
  \begin{tabular}{lcccccr}
    \textbf{source} & \textbf{type} & \textbf{language(s)} & \textbf{Domain} & \textbf{\#sentences} \\ \hline
    JW300 &  parallel & nl-srn  & religious  & 307,866 \\
    SIL & parallel & nl-srn & general & 2,927 \\
    Z-Library-1 & parallel & nl-srn & stories & 351 \\
    Z-Library-2 & parallel & nl-srn & stories & 163 \\
    Z-Library-3 & parallel & nl-srn & stories & 399 \\
    Naks Sranan fb & parallel & nl-srn & general & 62 \\
    Dutch DOJ  & parallel & nl-srn & legal & 220 \\
    SIL & monolingual & srn & stories & 6,572 \\\hline
    \textbf{total:} & & & & \textbf{318,560}
  \end{tabular}
  \caption{\small{Contents of SRNcorpus. The Summer Institute of Linguistics (SIL) sentences are scraped from an online Sranan dictionary. Z-library 1, 2 and 3 are authored by Tori di switi fu leisi / H.C. Tiendalli, Lafu tori / J. Redan and Dri Anansi tori / H.C. Tiendalli respectively. Naks Sranan is the facebook page of a cultural organisation and contains member introductions. Finally, the Dutch DOJ data contains warrants for arrest.}}
  \label{tab:SRNcorpus}
\end{table}

\hspace{-0.4cm}\textbf{Domain temperature sampling} Temperature sampling is often used to overcome size differences between language pairs by oversampling lower resource pairs. In contrast, we applied temperature sampling with the goal of reducing domain imbalance. We distinguish between religious vs non-religious data within SRNcorpus.\\
\vspace{-0.1cm}\textbf{Experimental setup} First we experiment with bilingual models (+ backtranslation). Then we investigate transfer learning, where we first train a parent model and then finetune on a child model. Finally we train multilingual models by sharing all parameters between languages and prepending a target token \cite{Johnson2017}. For all experiments we used the Transformer base architecture \cite{Vaswani2017}. We implemented all our models using JoeyNMT \cite{Kreutzer2019}. For evaluation we calculated BLEU scores using sacrebleu \cite{post-2018-call}.\\
\textbf{Results} See Table \ref{tab:results} for an overview of our results. 

\begin{table}[h]
    \small
    \centering
    \begin{tabular}{lcccc}
    \multicolumn{1}{l|}{\textbf{type}} &  $T$ & \multicolumn{1}{l|}{BPE} &  \multicolumn{1}{l}{\textbf{nl-srn}} & \multicolumn{1}{l}{\textbf{srn-nl}}        \\ \hline
    \multicolumn{1}{l|}{\texttt{bl}}     & - & \multicolumn{1}{l|}{1000}     & 22.06 & 15.04 \\ \hline
    \multicolumn{1}{l|}{\texttt{SRN}}    & 3 & \multicolumn{1}{l|}{3000}     & -     & 28.00 \\
    \multicolumn{1}{l|}{\texttt{SRN}}    & 2 & \multicolumn{1}{l|}{4000}     & 35.48 & -     \\
    \multicolumn{1}{l|}{\texttt{SRN}}    & 2 & \multicolumn{1}{l|}{3000}     & 36.78 & -     \\
    \multicolumn{1}{l|}{\texttt{SRN}}    & 2 & \multicolumn{1}{l|}{2000}     & 36.86 & -     \\
    \multicolumn{1}{l|}{\texttt{SRN}}    & 2 & \multicolumn{1}{l|}{1000}     & 37.48 & 27.47 \\
    \multicolumn{1}{l|}{\texttt{SRN}}    & 2 & \multicolumn{1}{l|}{500}      & 36.60 & -     \\
    \multicolumn{1}{l|}{\texttt{SRN}}    & 1 & \multicolumn{1}{l|}{1000}     & 32.92 & 24.93 \\
    \multicolumn{1}{l|}{\texttt{SRN}}    & 3 & \multicolumn{1}{l|}{1000}     & 36.99 & 28.47 \\ \hline
    \multicolumn{1}{l|}{\texttt{bt}}     & 2 & \multicolumn{1}{l|}{1000}     & \textbf{38.88} & - \\
    \multicolumn{1}{l|}{\texttt{bt$^\star$}}     & 2 & \multicolumn{1}{l|}{1000}     & 37.29 & -     \\
    \multicolumn{1}{l|}{\texttt{bt}}     & 3 & \multicolumn{1}{l|}{1000}     & -     & 27.41 \\
    \multicolumn{1}{l|}{\texttt{bt$^\star$}}      & 3 & \multicolumn{1}{l|}{1000}    & -     & 27.87 \\ \hline
    \multicolumn{1}{l|}{\texttt{tl}}     & 2 & \multicolumn{1}{l|}{32000}    & 37.33 & -     \\
    \multicolumn{1}{l|}{\texttt{tl}}     & 2 & \multicolumn{1}{l|}{16000}    & 37.49 & -     \\
    \multicolumn{1}{l|}{\texttt{tl}}     & 2 & \multicolumn{1}{l|}{8000}     & 38.85 & -      \\
    \multicolumn{1}{l|}{\texttt{tl}}     & 3 & \multicolumn{1}{l|}{32000}    & -     & \textbf{32.02} \\
    \multicolumn{1}{l|}{\texttt{tl}}     & 3 & \multicolumn{1}{l|}{16000}    & -     & 30.78 \\
    \multicolumn{1}{l|}{\texttt{tl}}     & 3 & \multicolumn{1}{l|}{8000}     & -     & 29.70 \\ \hline
    \multicolumn{1}{l|}{\texttt{ml}}     & 2 & \multicolumn{1}{l|}{32000}    & 30.04 & -     \\
    \multicolumn{1}{l|}{\texttt{ml}}     & 2 & \multicolumn{1}{l|}{16000}    & 29.88 & 23.02 \\
    \multicolumn{1}{l|}{\texttt{ml}}     & 2 & \multicolumn{1}{l|}{8000}     & 29.00 & -     \\
    \multicolumn{1}{l|}{\texttt{ml}}     & 3 & \multicolumn{1}{l|}{32000}    & -     & 21.71 \\
    \end{tabular}
    \caption{\small{BLEU scores reported on the non-religious test set of the SRNcorpus. $T$ is domain sampling temperature, BPE is the number of merge ops. \texttt{bl} is baseline trained using only religious JW300 data. \texttt{SRN} are bilingual models trained on SRNcorpus. \texttt{bt} are backtranslation experiments ($\star$ indicates inclusion of synthetic data). \texttt{tl} are transfer learning experiments. \texttt{ml} are multilingual experiments.}}
    \label{tab:results}
\end{table}

\hspace{-0.4cm}\textbf{Bilingual} We found that for both nl-srn and srn-nl, using SRNcorpus instead of JW300 increased BLEU scores by 10+ points. 1000 BPE merge operations produced best results, which is in line with \citep{gowda2020} who studied the effect of the number of BPE merge operations in relation to the parallel corpus size on translation performance. For nl-srn best results were obtained with $T=2$, whereas for srn-nl $T=3$ worked best.\\
\textbf{Backtranslation} For our backtranslation experiments we used the best performing bilingual srn-nl model (BPE=1000; $T=3$) to translate the monolingual data from SRNcorpus into nl. We then trained models in both directions using this synthetic generated data on top of our SRNcorpus. For BPE and $T$ we used the best values according to the bilingual models. We found a +1.4 BLEU increase for nl-srn, scoring highest overall for this direction, while our srn-nl model decreased slightly. In addition we experimented with applying domain temperature sampling to the synthetic parallel data as well. We found this to hurt performance for all translation directions.\\
\textbf{Transfer Learning} Since Sranan is an English based Creole language, the language pair nl-en is a natural choice as a parent model. After training the parent model until convergence, we initialized our child models with the resulting parameters. We increased the number of merge ops to accommodate for the increased vocabulary. For nl-srn, BPE=8k resulted in best translation quality, almost on par with the backtranslation experiment. For srn-nl, best results were achieved with BPE=32k, resulting in the highest BLEU score for this direction.\\
\textbf{Multilingual} For our multilingual models we used our SRNcorpus plus other data as described in the previous Section. We report scores for nl-srn (which are obtained by training a one-to-many model on nl-en, nl-pt and nl-srn) and srn-nl (which are obtained by training a many-to-one model on the reverse directions). Note that we applied language temperature sampling ($T=5$) to oversample nl-srn and nl-pt. We found that resulting models perform substantially worse compared to our other models (except the baseline).

\vspace{-0.2cm}\section{Conclusion}\vspace{-0.15cm}
In this study we have put NMT for Sranan on the map. We introduced the SRNcorpus and used it for various experiments in search for a performant general purpose machine translation for nl→srn and srn→nl. Our results demonstrate a strong baseline machine translation system for Sranantongo, which future work can build on.

\bibliography{references}

\end{document}